\documentclass[10pt, conference, compsocconf]{IEEEtran}

\usepackage{float}
\usepackage{multirow}
\usepackage{relsize}
\usepackage{cite,graphicx}
\usepackage{caption}
\usepackage{subcaption}

\usepackage{xcolor,colortbl}
\usepackage{amsmath,amssymb}
\usepackage{amsthm,amsmath}
\usepackage[ruled,linesnumbered,boxed,noend]{algorithm2e}

\newcommand{\ci}[3]{\ensuremath{I({#1},{#2} \mid #3})}

\newcommand{\cd}[3]{\ensuremath{ \lnot I({#1},{#2} \mid #3})}

\newcommand{\F}{\ensuremath{\mathcal{F}}}
\newcommand{\D}{\ensuremath{\mathcal{D}}}

\DeclareMathOperator{\val}{Val}
\DeclareMathOperator{\adj}{adj}

\newtheorem{definition}{Definition}

\newtheorem*{operation*}{Operation}
\newtheorem*{theorem*}{Theorem}
\newtheorem{example}{Example} 

\begin{document}
%
\title{Learning Markov networks with context-specific independences}

\author{\IEEEauthorblockN{ ~Alejandro~Edera, ~Federico~Schl\"uter, ~Facundo~Bromberg }
	\IEEEauthorblockA{Departamento de Sistemas de Informaci\'on, \\
	 Universidad Tecnol\'ogica Nacional, Facultad Regional Mendoza, Argentina. \\
	   \{aedera,federico.schluter,fbromberg\}@frm.utn.edu.ar}
}

\maketitle

\begin{abstract}
Learning the Markov network structure from data
is a problem that has received considerable attention 
in machine learning, and in many other application fields.
This work focuses on a particular approach for this purpose 
called \emph{independence-based} learning.
Such approach guarantees the learning of the correct structure efficiently, 
whenever data is sufficient for representing the underlying distribution.
However, an important issue of such approach is that the learned structures
are encoded in an undirected graph.
The problem with graphs is that they cannot encode 
some types of independence relations, such as the context-specific independences.
They are a particular case of conditional independences that is true only for a certain
assignment of its conditioning set, in contrast to conditional independences that must hold for all its assignments.
In this work we present CSPC, an independence-based algorithm 
for learning structures that encode context-specific independences,  
and encoding them in a log-linear model, instead of a graph.
The central idea of CSPC is combining the theoretical guarantees provided by the independence-based approach
with the benefits of representing complex structures by using features in a log-linear model.
We present experiments in a synthetic case, showing that CSPC is more accurate than
the state-of-the-art IB algorithms when the underlying distribution contains CSIs.
\end{abstract}


\IEEEpeerreviewmaketitle

\section{Introduction}

Nowadays, a powerful representation of joint probability distributions are Markov
networks. The structure of a Markov network can encode complex probabilistic
relationships among the variables of the domain, improving the efficiency in the procedures for
probabilistic inference. An important problem is learning the structure from samples drawn
from an unknown distribution. 
A number of alternative algorithms for this purpose have been developed in recent years. 
One approach is the \emph{independence-based} (IB) approach
\cite{Spirtes00,bromberg&margaritis09b,gandhi08,margaritisBromberg09,Bromberg&Schluter&Edera2011}.
Algorithms that follow this approach proceed by using statistical tests to learn a series of
conditional independences from data, encoding them in an undirected graph. 
An important advantage of this approach is that it provides theoretical
guarantees for learning the correct structure, together with the efficiency gained by using statistical tests. 
Other recent approaches
\cite{davis2010bottom,lowd2010learning,gogate2010learning,van2012markov}
proceed by inducing a set of features from data, instead of an undirected graph. 
The features are real-valued functions of partial variable assignments,
and using these functions it is possible to encode more complex structures than those encoded by graphs. 
Algorithms that follow this approach encode the structure in the features of a log-linear model.
Unfortunately, current algorithms based on learning features 
are not an efficient alternative, due to the multiple
user defined hyper-parameters, and the need of performing parameters learning. 
The parameters learning step is often intractable, requiring an iterative optimization 
that runs an inference step over the model at each iteration.

In many practical cases the underlying distribution of a problem
present \emph{context-specific independences} (CSIs)
\cite{DBLP:conf/uai/BoutilierFGK96}, that are conditional
independences that only hold for a certain assignment of the conditioning set, 
but not hold for the remaining assignments. 
In that case, encoding the structure in an undirected graph
leads to excessively dense graphs, obscuring the CSIs present in the distribution, 
and resulting in computationally more expensive computation of inference algorithms \cite{poole2003exploiting,fridman2003mixed}. 
For this reason, encoding the CSIs in a log-linear model
does not obscure them, achieving sparser models
and therefore significant improvements in time, space and
sample complexities \cite{benedek2008mixed,fierens-csiidrpmaiioteogs,wexler-ifmm,gogate2010learning}.

This work presents CSPC, an independence-based algorithm 
for learning a set of features instead of a graph, in order to encode CSIs.
The algorithm is designed as an adaptation for this purpose of the well-known PC algorithm
\cite{spirtes1991algorithm}.  
CSPC proceeds by first generating an initial set of features from the dataset, 
and then searches over the space of possible contexts for learning the CSIs present in the underlying distribution. 
For each context the algorithm elicits a set of CSIs using statistical tests, 
and generalizes the current set of features in order to encode the elicited CSIs.
The central idea of CSPC is combining the theoretical guarantees provided by the independence-based approach
with the benefits of representing complex structures by using features.
To our knowledge, the only algorithm near to CSPC is the LEM algorithm \cite{gogate2010learning}, since 
it uses statistical tests to learn CSIs. 
However, LEM restricts the attention to learning distributions that can be represented by
decomposable Markov networks. For the latter, we omit it as competitor in our experiments.

We conducted an empirical evaluation on synthetic data generated from
known distributions that contains CSIs. In our experiments we prove
that CSPC is significantly more accurate than the state-of-the-art IB algorithms when the underlying distribution contains CSIs.

\section{Background}
This section reviews the basics about Markov networks representation,
the concept of CSIs, and the IB approach for learning Markov networks.

\subsection{Markov networks}

A Markov network over a domain $X$ of $n$ random variables $X_0 \ldots X_{n-1}$
is represented by an undirected graph $G$ with $n$ nodes
and a set of numerical parameters
$\theta\in\mathbb{R}$.
This representation can be used to factorize the distribution
with the Hammersley-Clifford theorem \cite{Hammersley_Clifford_1971}, 
by using the completely connected sub-graphs
of \(G\) (a.k.a., \emph{cliques}) into a set of \emph{potential
  functions} \(\{\phi_C(X_C)~ \colon~ C\in cliques(G)\}\) of lower dimension than
\(p(X)\), parameterized by \(\theta\), as follows:
  \begin{align} \label{eq:Gibbs}
    p(X=x) &= \frac{1}{Z} \prod_{C \in cliques(G)} \phi_C(x_C) ,
  \end{align}

  where $x$ is a complete assignment of the domain $X$, $x_C$ is the
  projection of the assignment \(x\) over the variables of the $C$th
  clique, and $Z$ is a normalization constant.
  An often used alternative representation is a \emph{log-linear} model,
  with each clique potential represented as an exponentiated weighted
  sum of features of the assignment, as follows:

\begin{align} \label{eq:log-linear-indicator}
  p(X=x) &= \frac{1}{Z} \exp \left\{~ \sum_j \theta_{j} f_j(x) \right\},
\end{align}

where each feature $f_j$ is a partial assignment over a subset of the
domain $V(f_j)$.
%
%
Given an assignment $x$, a feature $f_j$ is said to be satisfied iff
for each single variable $X_a=x_a \in f_j$ it also holds that $x_a \in
x$ \cite{gogate2010learning}. One can associate a indicator function
to \(f_j\) and an assignment \(x\) by associating a value \(1\) when
\(f_j\) is satisfied in \(x\), or \(0\) otherwise.

A Markov network can be induced from a log-linear model 
by adding an edge in the graph between every pair of variables
$X_a,X_b$ that appear together in some subset of a feature $f_j$, that
is $\{X_a,X_b\} \subseteq V(f_j)$.  Then, the clique potentials are
constructed from the log-linear features in the obvious way
\cite{Lee+al:NIPS06}. 

\begin{example} \label{example:1}
Figure~\ref{fig:fully} shows the features of a log-linear model 
over $n=3$ binary variables $X_f$, $X_a$ and $X_b$,
and its respective induced graph.

\begin{figure}[ht]

  \begin{minipage}{0.35\textwidth}
  \centering 
  \scriptsize
    \begin{tabular}{ c c c c }
	\(f_1\) & \hspace{-0.35cm}\((X_a=0\) & \hspace{-0.35cm}\(X_b=0\) & \hspace{-0.35cm}\(X_f=0)\) \\
       \(f_2\) & \hspace{-0.35cm}\((X_a=1\) & \hspace{-0.35cm}\(X_b=0\) & \hspace{-0.35cm}\(X_f=0)\) \\
       \(f_3\) & \hspace{-0.35cm}\((X_a=0\) & \hspace{-0.35cm}\(X_b=1\) & \hspace{-0.35cm}\(X_f=0)\) \\
       \(f_4\) & \hspace{-0.35cm}\((X_a=1\) & \hspace{-0.35cm}\(X_b=1\) & \hspace{-0.35cm}\(X_f=0)\) \\
       \(f_5\) &  \hspace{-0.35cm}\((X_a=0\) & \hspace{-0.35cm}\(X_f=1)\)&          \\
       \(f_6\) & \hspace{-0.35cm}\((X_a=1\) & \hspace{-0.35cm}\(X_f=1)\)& \\
       \(f_7\) &  \hspace{-0.35cm}\((X_b=0\) & \hspace{-0.35cm}\(X_f=1)\)&          \\
       \(f_8\) & \hspace{-0.35cm}\((X_b=1\) & \hspace{-0.35cm}\(X_f=1)\)& \\
     \end{tabular} 
     \textbf{$\mathlarger{\mathlarger{\mathlarger{\mathlarger{\Rightarrow}}}}$}
  \end{minipage}\hspace{-.7cm}
     \begin{minipage}{2.2cm}
      \includegraphics[width=2.2cm]{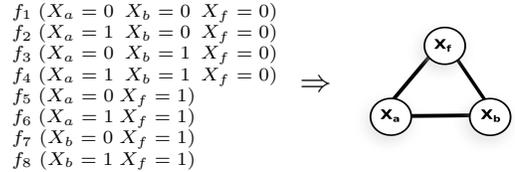}  
     \end{minipage}

  \caption{An example of an induced graph from a set of features.}
  \label{fig:fully}
\end{figure}
\end{example}

\subsection{Context-specific independences}

The CSIs are a finer-grained type of independences.
These independences are similar to conditional independences,
but hold for a specific assignment of the conditioning set, called
the \emph{context} of the independence. 
Formally, we define a CSI as follows:
\begin{definition}[Context-specific independence
  \cite{DBLP:conf/uai/BoutilierFGK96}]
  Let $X_a , X_b \in X$ be two random variables, $X_U,X_W \subseteq X \setminus\{X_a,X_b\}$
  be pairwise disjoint sets of variables that does not contain $X_a,X_b$;
  and $x_W$ some assignment of $X_W$.
  We say that variables \(X_a\) and \(X_b\) are
  \emph{contextually independent} given \(X_U\) and
  a context $X_W=x_W$, denoted $\ci{X_a}{X_b}{X_U,x_W}$, iff
  \begin{align} \label{eq:csi}
    p(X_a \vert  X_b, X_U, x_W) = p(X_a \vert X_U, x_W),
  \end{align}
  whenever \(p(X_b, X_U, x_W) > 0\).
\end{definition}

\begin{example}  \label{example:2}
Figure~\ref{fig:csi}(a) shows the graph of Example~\ref{example:1},
induced from a log-linear model.
Notice that the features of Example~\ref{example:1} encode 
the CSI $\ci{X_a}{X_b}{X_f=1}$, but it is obscured in the graph. 
Alternatively, such CSI can be graphically represented 
if we use two graphs, one for each value of $X_f$.
For this, Figure~\ref{fig:csi}(b)  shows the graph induced from the features with $X_f=1$
which encodes $\ci{X_a}{X_b}{X_f=1}$, and  
Figure~\ref{fig:csi}(c)  shows the graph induced from the features with $X_f=0$
which encodes $\cd{X_a}{X_b}{X_f=0}$. 
In these figures, gray nodes correspond to an assignment of a variable. 
\begin{figure}[h]
  \centering
    \includegraphics[width=0.45\textwidth]{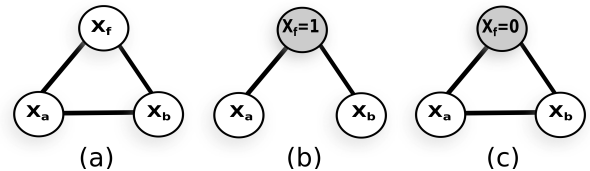} 
  \caption{(a) The graph induced from the features in Example~\ref{example:1}.
(b) graph induced from the features with $X_f=1$ in Example~\ref{example:1}.
(c) graph induced from the features with $X_f=0$ in Example~\ref{example:1}.
Gray nodes correspond to an assignment of a variable.
}
  \label{fig:csi}
\end{figure}
\end{example}

Notice that the graph in Figure~\ref{fig:csi}(a) 
cannot encode the CSI $\ci{X_a}{X_b}{X_f=1}$, 
because it occurs only for a specific context and is
absent in all the others.  
This is because the edges connect pairs of
variables that are conditionally dependent even for a single choice of
values of the other variables.  Since a CSI is defined for a specific
context, a set of CSIs cannot be encoded all together in a single undirected graph \cite{koller09}.
Nonetheless, both structures are encoded in the set of features of the example.

\subsection{Independence-based approach for structure learning}
\label{sec:IB-approach}

The task of IB~algorithms is learning a graph that encodes
the independences from i.i.d. samples $\D = \{x^1, \ldots, x^D\}$ of an
unknown underlying distribution $p(X)$ \cite{Spirtes00}. 
For that, these algorithms perform a succession of statistical independence tests over \D\ to
determine the truth value of a conditional independence (e.g. Mutual
Information \cite{cover&tomas91}, Pearson's $\chi^2$ and $\mathcal{G}^2$
\cite{AGRESTI02}), discarding all graphs that are inconsistent with
the test. The decision of what test to perform is based on the
independences learned so far, and varying with each specific algorithm. 

A key advantage of these algorithms is that they guarantees to learn the correct underlying structure
under three assumptions: (\textit{i}) the
underlying distribution is \emph{graph-isomorph}, that is, the
independences in \(p(X)\) can be encoded by a graph; (\textit{ii}) the
underlying distribution is positive, that is $p(x)>0$ for all $X=x$;
and (\textit{iii}) the outcomes of statistical independence tests are
correct, that is the independences learned are a subset of the
independences present in \(p(X)\).  
Another advantage of using IB algorithms is its computational efficiency, due to its polynomial
running time \cite{Spirtes00}, and also due to the avoiding of the
need of performing parameters learning.  The efficiency is gained
because the computational cost of a test is proportional to the number
of rows in $\mathcal{D}$, and the number of variables involved in the tests.

Perhaps the best known algorithm that follows this approach is PC
\cite{spirtes1991algorithm}, which was created for learning the
structure of Bayesian networks.  PC is correct under the assumptions
described above, but when the tests are not correct produce errors in
removing edges, because the algorithm only tests for independence among
two variables conditioning in subsets of the adjacencies of one of these variables.  
For learning Markov networks, the first algorithm that follows the IB approach is
GSMN \cite{bromberg2009efficient}, an efficient algorithm that
computes only $O(n^2)$ tests, constructing the structure by learning the
adjacencies of each variable; using the \emph{Grow-Shrink}
algorithm \cite{margaritis00}.  A more recent algorithm that improves
over GSMN is IBMAP-HC \cite{bromberg2011independence}, which learns the
structure by performing a hill-climbing search
over the space of graphs looking for the one which maximizes the
IB-score, a score of the posterior probabilities of graphs $p(G \mid \mathcal{D})$.  
The hill-climbing search starts from the empty structure, 
adding edges until reaching a local maxima of $p(G \mid \mathcal{D})$.
IBMAP-HC relaxes the assumption about the correctness of the
statistical tests, improving over GSMN in sample complexity by reducing the cascade effect of incorrect tests.

\section{Context-Specific Parent and Children algorithm}
\label{sec:cpc}

This section presents the CSPC (Context-specific Parent and Children)
algorithm for learning Markov networks structures that encodes the
CSIs present in data. CSPC encodes the CSIs by generalizing
iteratively a set of features. 
For this, CSPC decomposes the search space of CSIs in two nested spaces:
the space of the possible contexts, and for each context the space of
all its possible CSIs.  First, CSPC generates an initial set of
features, and then searches over both spaces with two nested loops: an
outer loop that explores the space of the contexts; and for each one,
an inner loop that elicits from data a set of CSIs by using statistical
tests, and generalizes the features according to the elicited CSIs.

The three key elements of CSPC are: \textit{A)} the generation of the
initial features, \textit{B)} the elicitation of CSIs from data, and
\textit{C)} the features generalization for encoding the elicited CSIs.

\subsection{The generation of the initial features}

An initial set of features \F\ must be generated as a
starting point of the whole algorithm. One alternative is to generate
the features that correspond with the initial fully connected graph
in a similar fashion than PC, that is adding a feature 
for each possible complete assignment \(x\) of the variable \(X\). 
In this case, the size of such initial set is
exponential with respect to the number of variables. For this reason,
CSPC uses a more optimal initial set of features, adding one feature
for each unique example in $\mathcal{D}$.  This is an often used alternative
\cite{davis2010bottom,van2012markov}, because it guarantees that the generalized
features at the end of the algorithm match at least one training example.

\subsection{Eliciting context-specific independences}

For discovering all the CSIs present in the data, CSPC explores the
set of complete contexts \(x\) found in the dataset, that is, 
one for each unique training example.
Given a context \(x\in\mathcal{D}\) and a set of features \(\F\), CSPC
decides what CSIs to elicit in a similar fashion than PC.  
For each variable \(X_a\) a Markov network 
is induced from the subset of features that satisfies with the
context \(x_{X\setminus X_a}\). 
Then, for each \(X_b\) adjacent to \(X_a\) in the induced graph
 a subset \(X_{W}\) of the adjacencies is taken
 in order to define the conditional independence \(\ci{X_a}{X_b}{X_W}\). 
From such independence, a CSI is obtained by
contextualizing the conditioning set \(X_W\) using the context \(x\),
that is \(\ci{X_a}{X_b}{X_W=x_w}\). 
Finally, if the CSI is present in data then it is encoded in the
current set of features.

For eliciting the CSIs in data, we propose a straightforward
adaptation of a traditional (non-contextualized) independence
test. A similar adaptation is proposed in \cite{gogate2010learning}. 
The central idea of the adaptation is that an arbitrary CSI
$\ci{X_a}{X_b}{X_U,x_W}$ can be seen as a conditional independence
$\ci{X_a}{X_b}{X_U}$ in the conditional distribution $p(X \setminus
\{X_W\} \mid x_W)$. In this way, the CSI can be tested by using a
non-contextualized test over a sample drawn from the conditional
distribution $p(X \setminus \{X_W\} \mid x_W)$. In practice, such
sample can be obtained from $\D$ as \(\{x^j\in\D \colon x^j_W =
x_W\}\), namely, the subset of datapoints where $X_W=x_W$.

\subsection{Features generalization}

The generalization of features is used by CSPC to encode the CSIs that
are present in data. A specific CSI $\ci{X_a}{X_b}{x_W}$ can be
encoded in the features \F\ of a log-linear of \(p(X)\) by factorizing
those features that correspond with the conditional distribution $p(X
\setminus \{X_W\} \mid x_W)$. Such factorization is done by using a
recently proposed adaptation of the well known Hammersley-Clifford
theorem for CSIs called the Context-Specific Hammersley-Clifford theorem \cite{edera2013markov}. 
The features that correspond with $p(X
\setminus \{X_W\} \mid x_W)$ are the subset of features in $\F$ that satisfy
the context \(x_W\), denoted by \(\F[x_W]\subseteq \F\). In this way,
given the CSI $\ci{X_a}{X_b}{x_W}$ the features \(\F[x_W]\) are
factorized into two new sets of features: $\F'[x_W]$, obtained from
\(\F[x_W]\) but removing the variable \(X_a\); and $\F''[x_W]$, 
obtained from \(\F[x_W]\) but removing the variable \(X_b\). Formally, for all
\(f'_j\in\F'[x_W]\), \(\{X_a\}\notin V(f'_j)\); and for all
\(f''_j\in\F''[x_W]\), \(\{X_b\}\notin V(f''_j)\).

 \begin{example} 
 Figure~\ref{fig:fea.a} shows an initial set of features  $\F$
 to be generalized in order to encode the CSI \(\ci{X_a}{X_b}{X_f=1}\).
 The generalization consists in factorizing the features \(\F[X_f=1]\), 
 that is the set of features that are satisfied with the context $X_f=1$ (Figure~\ref{fig:fea.b}).
 The factorization of these features results in two new sets of features: $\F'[X_f=1]$ and $\F''[X_f=1]$,
 shown in Figure~\ref{fig:fea.c}. 
 The features in $\F'[X_f=1]$ are obtained from \(\F[X_f=1]\) but removing \(X_b\),
 and the features in $\F''[X_f=1]$ are obtained from \(\F[X_f=1]\) but removing \(X_a\)
 Finally, the set of features which correctly encodes the CSI \(\ci{X_a}{X_b}{X_f=1}\)
 are shown in Figure~\ref{fig:fea.d}. 
 Notice that the features in Figure~\ref{fig:fea.d} are the same set of features shown in Example~\ref{example:1}.
 
 \begin{figure}[!th] 
   \begin{subfigure}{0.17\textwidth}
     \centering
     \scriptsize
     \begin{tabular}{ c c c c }
       \(f_1\) & \hspace{-0.35cm}\((X_a=0\)  & \hspace{-0.35cm}\(X_b=0\) & \hspace{-0.35cm}\(X_f=0)\) \\
       \(f_2\) & \hspace{-0.35cm}\((X_a=1\) & \hspace{-0.35cm}\(X_b=0\) & \hspace{-0.35cm}\(X_f=0)\) \\
       \(f_3\) & \hspace{-0.35cm}\((X_a=0\) & \hspace{-0.35cm}\(X_b=1\) & \hspace{-0.35cm}\(X_f=0)\) \\
       \(f_4\) & \hspace{-0.35cm}\((X_a=1\) & \hspace{-0.35cm}\(X_b=1\) & \hspace{-0.35cm}\(X_f=0)\) \\
       \rowcolor{lightgray}
       \(f_5\) & \hspace{-0.35cm}\((X_a=0\) & \hspace{-0.35cm}\(X_b=0\) & \hspace{-0.35cm}\(X_f=1)\)          \\
       \rowcolor{lightgray}
       \(f_6\) & \hspace{-0.35cm}\((X_a=1\) & \hspace{-0.35cm}\(X_b=0\) & \hspace{-0.35cm}\(X_f=1)\)          \\
       \rowcolor{lightgray}
       \(f_7\) & \hspace{-0.35cm}\((X_a=0\) & \hspace{-0.35cm}\(X_b=1\) & \hspace{-0.35cm}\(X_f=1)\) \\
       \rowcolor{lightgray}
       \(f_8\) & \hspace{-0.35cm}\((X_a=1\) & \hspace{-0.35cm}\(X_b=1\) & \hspace{-0.35cm}\(X_f=1)\) \\
     \end{tabular}
     \caption{\small Initial set of features \label{fig:fea.a}}
   \end{subfigure}
   \quad \quad %
   \begin{subfigure}{0.2\textwidth}
     \scriptsize
     \begin{equation*}
     \begin{aligned}
       \F&[X_f=1] = \{\\
       &f_5(X_a=0 X_b=0 X_f=1),\\
       &f_6(X_a=1 X_b=0 X_f=1),\\
       &f_7(X_a=0 X_b=1 X_f=1),\\
       &f_8(X_a=1 X_b=1 X_f=1)\}\\
       \end{aligned}
     \end{equation*}
     \caption{The features that satisfy with
       the context \(X_f=1\) \label{fig:fea.b}}
   \end{subfigure}    
   \newline
   \begin{subfigure}{0.2\textwidth}
     \scriptsize
     \begin{equation*}
     \begin{aligned}
       \F'&[X_f=1] = \{\\
       &f'_5(X_b=0 X_f=1),\\
       &f'_6(X_b=0 X_f=1),\\
       &f'_7(X_b=1 X_f=1),\\
       &f'_8(X_b=1 X_f=1)\}\\
       \\
       \F''&[X_f=1] = \{\\
       &f''_5(X_a=0 X_f=1),\\
       &f''_6(X_a=1 X_f=1),\\
       &f''_7(X_a=0 X_f=1),\\
       &f''_8(X_a=1 X_f=1)\}\\
       \end{aligned}
     \end{equation*}
     \caption{ \small Factorization of the features \(\F[X_f=1]\) \label{fig:fea.c}}
   \end{subfigure}    
   \quad \quad %
   \begin{subfigure}{0.2\textwidth}
     \centering
     \scriptsize
     \begin{tabular}{c c c c }
       & & & \\
       & & & \\
       & & & \\
       & & & \\
       & & & \\
       & & & \\
       & & & \\
       & & & \\
       & & & \\
     \(f_1\) & \hspace{-0.35cm}\((X_a=0\) & \hspace{-0.35cm}\(X_b=0\) & \hspace{-0.35cm}\(X_f=0)\) \\
       \(f_2\) & \hspace{-0.35cm}\((X_a=1\) & \hspace{-0.35cm}\(X_b=0\) & \hspace{-0.35cm}\(X_f=0)\) \\
       \(f_3\) & \hspace{-0.35cm}\((X_a=0\) & \hspace{-0.35cm}\(X_b=1\) & \hspace{-0.35cm}\(X_f=0)\) \\
       \(f_4\) & \hspace{-0.35cm}\((X_a=1\) & \hspace{-0.35cm}\(X_b=1\) & \hspace{-0.35cm}\(X_f=0)\) \\
       \(f'_5\) &  \hspace{-0.35cm}\((X_a=0\) & \hspace{-0.35cm}\(X_f=1)\)&          \\
       \(f'_7\) & \hspace{-0.35cm}\((X_a=1\) & \hspace{-0.35cm}\(X_f=1)\)& \\
       \(f''_5\) &  \hspace{-0.35cm}\((X_b=0\) & \hspace{-0.35cm}\(X_f=1)\)&          \\
       \(f''_7\) & \hspace{-0.35cm}\((X_b=1\) & \hspace{-0.35cm}\(X_f=1)\)& \\
       \end{tabular}
     \caption{ \small Features generalized encoding
       $\ci{X_a}{X_b}{x_W}$ \label{fig:fea.d}}
   \end{subfigure}    
   \caption{ Example of feature factorization according to CSIs.}
 \end{figure}

\end{example}

\subsection{Overview}

This section presents an explanation of CSPC that puts all the pieces together.
The pseudocode is shown in Algorithm~\ref{alg:cspc:outer}. 
As input, the algorithm receives the set of domain variables \(X\), 
and a dataset \(\D\). The algorithm starts by generating the initial set of features. Then, the space of
the contexts is explored. For each context, the current set of
features is generalized by using a generalization of the PC algorithm
as a subroutine.  This subroutine, described in
Algorithm~\ref{alg:cspc:inner}, consists in the elicitation of CSIs
and the features generalization steps.  As input, this subroutine
receives the current set of features \F, the context \(x\), the set of
domain variables \(X\), and the dataset \D.  At the end, the features of a log-linear model are returned.

\begin{algorithm}[!h]\small
  \KwIn{domain variables \(X\), dataset \(\D\)}
  \KwOut{features \F\  generalized according to the CSIs learned}

  \BlankLine

  $\F \leftarrow$ Generate one feature for each unique example in $\mathcal{D}$
  \(x\in\val(X)\)\nllabel{alg:cpc:make-features}

  \ForEach{context \(x\in\val(X)\)}
  { \nllabel{alg:cpc:contexts}
    $\F \leftarrow$ PC (\F, \(x\), \(X\), \D)
  }
  
  Add atomic feature for each variable to $\F$
  
  \Return{$\F$}  \nllabel{alg:cpc:return}

  \caption{Context space exploration}
  \label{alg:cspc:outer}
\end{algorithm}

In Algorithm~\ref{alg:cspc:inner}, the step of elicitation of CSIs
follows the same strategy than PC, trying to find the independences on
the smallest number of variables in the conditioning set.  For this, the conditioning
set for each variable \(X_a\) consists on subset of size $k$ of the
adjacencies \(\adj(a)\),
terminating when for all subsets $W$, \(|W|\) is
smaller than $k$. This is a strategy for avoiding the effect of incorrect tests, 
because in the practice the quality of statistical tests decreases exponentially with the number of
variables that are involved \cite{AGRESTI02}. 

\begin{algorithm}[!h]\small
  \KwIn{features \F, context \(x\), domain variables \(X\), dataset \(\D\)}
  \KwOut{generalized features \F}

  \BlankLine

  $k \leftarrow 0$    \nllabel{alg:cpc:k0}

  \Repeat{\(\vert \adj(a) \setminus \{X_b\} \vert < k\) }
  {\nllabel{alg:cspc:subsets}

    \ForEach{$X_a\in X$} {
      \(\adj(a)\) $\leftarrow$ compute adjacencies from features that satisfy \(x_{X\setminus X_a}\) \nllabel{alg:cpc:adjs}

      \ForEach {$X_b \in  \adj(a)$ } { \nllabel{alg:cpc:edges}
          \ForEach{$W$ subset of adj($a$) $\setminus \{X_b\}$ s.t. $\vert W\vert = k$}
          {\nllabel{alg:cpc:context}

            \If{$\ci{X_a}{X_b}{x_W}$ is true} { \nllabel{alg:cpc:test}
	      $\F \leftarrow $ Generalize \(\F\) for the CSI $\ci{X_a}{X_b}{x_W}$
            }
          }
        }

      }
      $k \leftarrow k + 1$ \nllabel{alg:cpc:incremental}
    }
  \Return{$\F$}  \nllabel{alg:cpc:return2}

  \caption{PC extended for features}
  \label{alg:cspc:inner}
\end{algorithm}

In Algorithm~\ref{alg:cspc:generalization}, the step of features
generalization is made once the CSI has been elicited.  In such
step, for the input CSI $\ci{X_a}{X_b}{x_W}$ the current set of features \(\F\) is partitioned in two
sets: the set \(\F'\) that are the features that satisfy with $x_W$, 
and the set \(\F''\) that are the features that does not satisfy 
with $x_W$.  Then, the satisfied features are factorized
according to the Context-Specific Hammersley-Clifford theorem. 
Finally, a new set of features \(\F\) is
defined by joining \(\F'\) and \(\F''\).

\begin{algorithm}[!h]\small
  \KwIn{features \F, a CSI $\ci{X_a}{X_b}{x_W}$}
  \KwOut{generalized features \F}

  \(\F' \leftarrow\) $\F[x_W]$

  \(\F'' \leftarrow\) $\F\setminus\F'$

  $\F' \leftarrow $ factorize $\F'$ according to the Context-Specific Hammersley-Clifford 

  \Return{\(\F'\cup\F''\)}

  \caption{Feature generalization}
  \label{alg:cspc:generalization}
\end{algorithm}

\section{Experimental evaluation}
\label{sec:experiments}

To allow a proper experimental design with a range of well-understood
conditions, we evaluated our approach on artificially generated
datasets. For testing the effectiveness of our approach, 
we propose a specific class of deterministic models
which presents a controlled number of CSIs.
The evaluation consists in two parts. 
In the first part we show the potential of improvements that can be obtained
in our experiment.
In the second part we compare CSPC to two state-of-the-art IB algorithms: GSMN
\cite{bromberg2009efficient} and IBMAP-HC
\cite{bromberg2011independence}. Additionally we also compare with PC
\cite{spirtes1991algorithm} in order to highlight the improvements
resulting from contextualizing it. 
Since PC was originally designed for learning Bayesian networks (directed graphs),
we use PC omitting the step of edges orientation \cite{kalisch2007estimating}.

\subsection{Datasets} \label{sec:datasets}

We generated artificial data through Gibbs sampling 
on a class of models similar to Example~\ref{example:2},
generalized to distributions with $n$ discrete binary variables.
We chose such models since they are a representative case of
a distribution with a controlled number of CSIs. 
The aim is to demonstrate that learning such CSIs represents an important improvement
in the quality of learned distributions,
when compared with the alternative representation in graphs. 
In this scenario, the IB algorithms lead to excessively dense graphs (the fully connected ones),
which obscures the underlying CSIs.
We considered models with $n \in \{6,7,8\}$ variables and maximum cliques of the same size.
Since the complexity of structure learning grows exponentially with the size of its maximum
clique (a.k.a. treewidth), in the literature the algorithms are typically tested on models with
maximum cliques of size at most $6$ \cite{gogate2010learning,Lee+al:NIPS06}. 

For each $n$, the underlying structure is a fully connected graph with $(n-1)$ nodes,
plus a flag node $X_f$.
In this model, all pairs between the variables  
\(X \setminus \{X_f\}\) are context-specific independent, given the context $X_f=1$.
Instead, when $X_f=0$ the variables remain dependent. 
In this way, for $n$ variables the underlying structure contains 
$\frac{(n-1)\times(n-2)}{2}$ contextual independences 
in the form $\ci{X_a}{X_b}{X_f=1}$, for all $X_a,X_b \neq X_f \in X$.
Given such structure, we defined a log-linear model 
that contains two sets of features: \textit{i)} a set of \emph{pairwise} features
which encodes the dependence between $X_f$ and the rest of variables $X\setminus \{X_f\}$,
and \textit{ii)} a set of \emph{triplet} features over the variables $X_a,X_b,X_f$.
For the resulting features we generated $10$ different models, varying in its numerical parameters.
Such parameters were generated to satisfy the log-odds ratio, 
in order to set strong dependencies in the model \cite{AGRESTI02,gandhi08,margaritisBromberg09,Bromberg&Schluter&Edera2011}.
In this way, the parameters of the pairwise features $X_a,X_b$ 
were forced to satisfy the following ratio:
$\varepsilon = \log \left(
  \frac{w_0 \phi(X_a=0,X_f=0) w_1 \phi(X_a=1,X_f=1)}{w_2 \phi(X_a=0,X_f=1)
    w_3 \phi(X_a=1,X_f=0)} \right), \forall X_a \in X \setminus \{X_f\}$,
where $w_0,w_2$ are symmetric to $w_1,w_3$, respectively ($w_0=w_1$ and $w_2=w_3$).
Since this ratio has $2$ unknowns we choose $w_2$ sampled from $\mathcal{N}(0.5;0.001)$, and $w_0$ is solved.  
The parameters for the triplet features were forced to satisfy the CSI $\ci{X_a}{X_b}{X_f=1}$.
When $X_f=0$ the parameters were generated using the same procedure used for the pairwise features.
When $X_f=1$ the parameters were forced to satisfy the following factorization:
$\phi(X_a,X_b,X_f=1) = w_0 \phi(X_a=0,X_f=1) \cdot w_1 \phi(X_a=1,X_f=1)$,
where $w_0$ and $w_1$ are the same than the pairwise features already defined.
In our experiments we set $\varepsilon = 1.0$.  
The datasets were generated by sampling from the log-linear models
using Rao-Blackwellized Gibbs sampler \footnote{Gibbs sampler is
  available in the open-source Libra toolkit
  http://libra.cs.uoregon.edu/} with 10 chains, 100 burn-in and 1000
sampling.

\subsection{Methodology}


We used the synthetic datasets explained above to learn the structure and parameters for all the algorithms.
Our synthetic data, together with an executable version of CSPC 
and the competitors is publicly available\footnote{http://dharma.frm.utn.edu.ar/papers/cspc}.
For a fair comparison, we use Pearson's $\chi^2$ as the statistical independent test for all the algorithms, 
with a significance level of $\alpha=0.05$. 
The IBMAP-HC algorithm alternatively only works by using the Bayesian test of Margaritis \cite{margaritis00}.
For each particular dataset we evaluated the algorithms on training set sizes varying from $500$ to $40000$, 
in order to obtain a number of samples sufficient for satisfying the CSIs of the underlying distribution proposed.

We report the quality of learned models using the Kullback-Leibler divergence (\emph{KL}) \cite{kullback1951information}. 
The KL is defined as $KL(p \mid\mid q)=\sum_x p(x) ln \frac{p(x)}{q(x)}$, 
measuring the information lost when the learned distribution \(q(X)\) is used to approximate the
underlying distribution \(p(X)\). KL is equal to zero when \(p(X) =
q(X)\). The better the learned models, the lower the values of the KL measure. 
Since these algorithms only learn the structure of a Markov
network, the complete distribution is obtained by learning its
numerical parameters. For the case of IB algorithms, the features are induced from the maximum cliques of the graph learned. 
For learning its parameters we computed the pseudo-loglikelihood 
using the available version in the Libra toolkit. We use pseudo-loglikelihood without
regularization 
to avoid sparsity in the final model, because we are interested in measuring the quality of the structure learning step.

\subsection{Results}

\begin{figure*}[t] 
  \centering
    \includegraphics[width=0.32\textwidth]{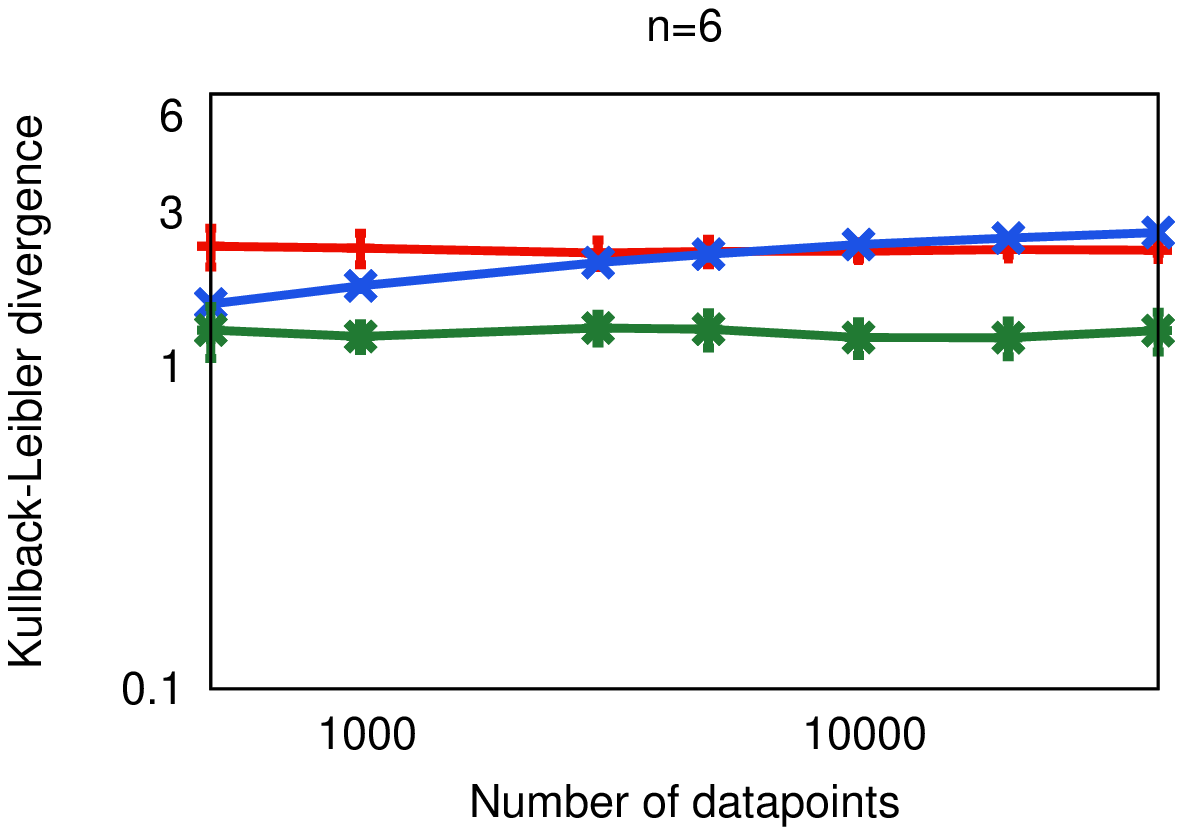}
    \includegraphics[width=0.32\textwidth]{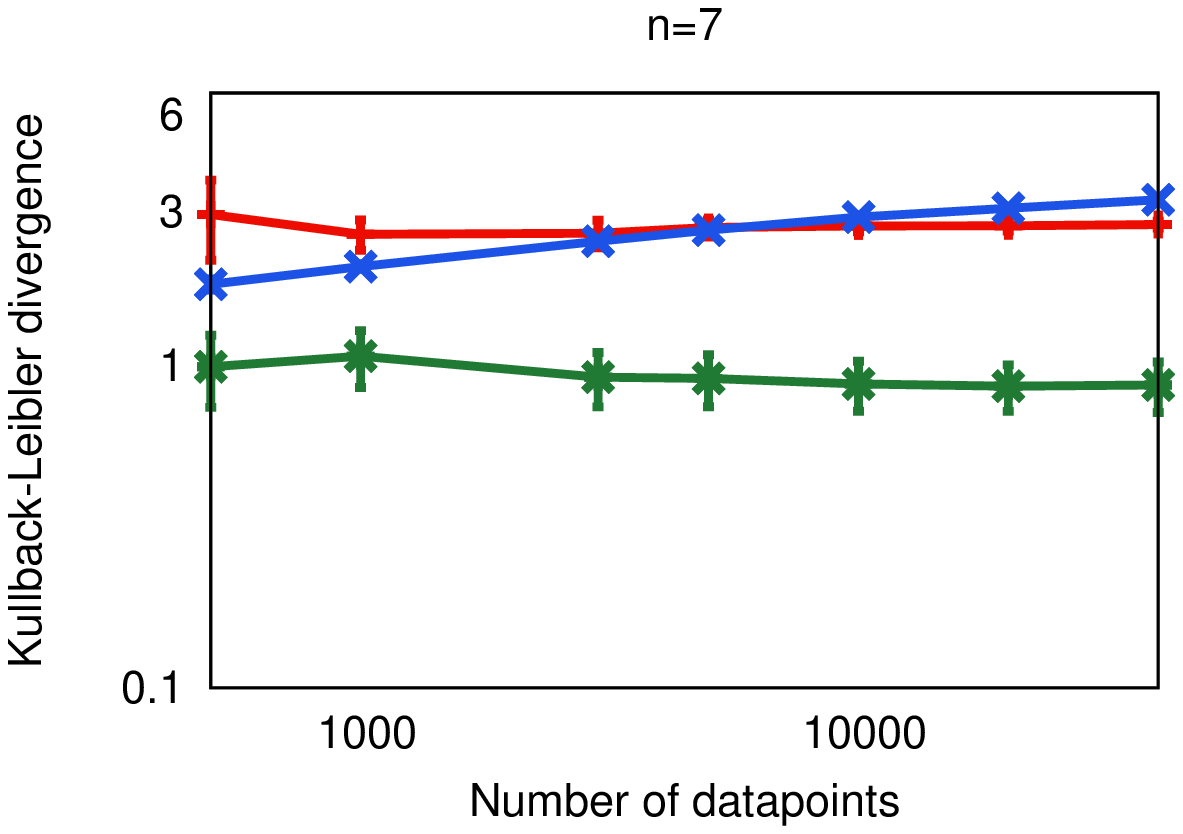} 
     \includegraphics[width=0.32\textwidth]{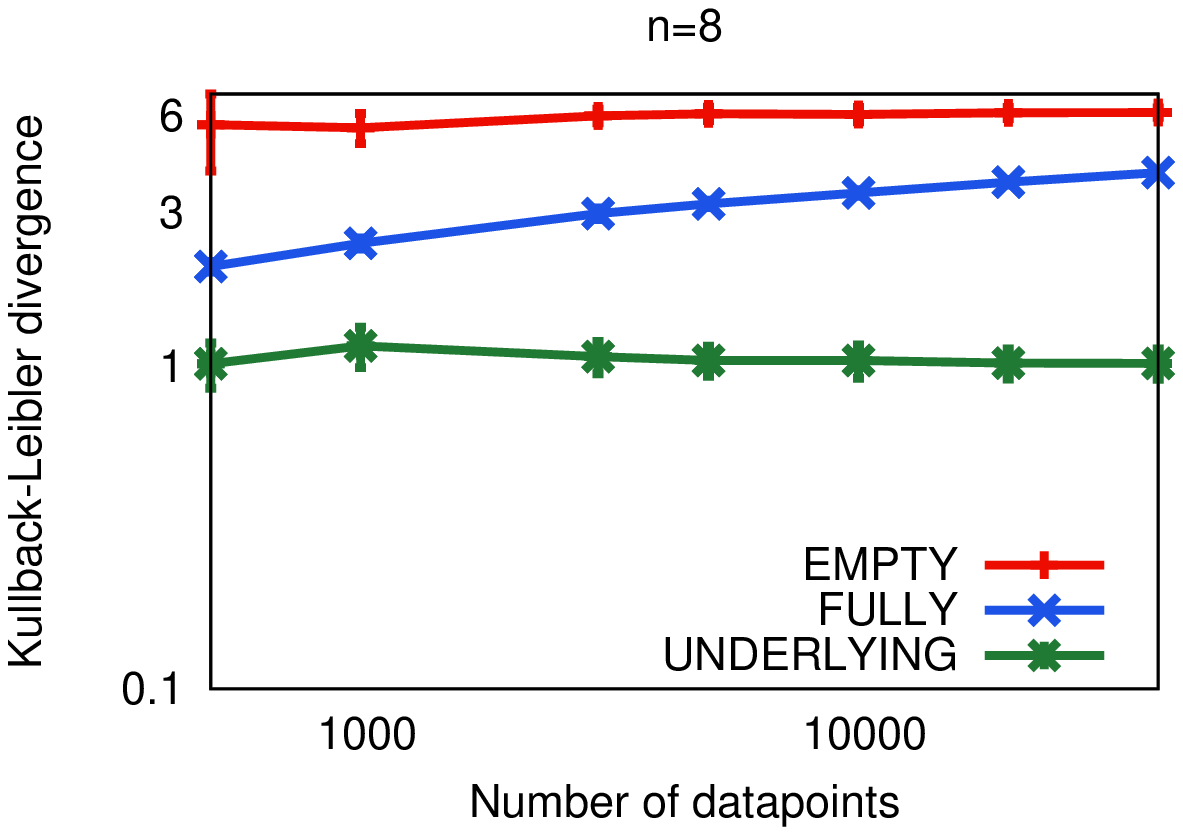}        
  \caption{
     Potential improvements in KL obtained by learning parameters for the underlying structure, 
     the fully and the empty structures.
     Average and standard deviation over ten repetitions for increasing number of datapoints 
     in the training set for domain sizes $6$ (left), $7$ (center) and $8$ (right). 
  }
  \label{fig:KLUnderlying}
\end{figure*}

Our first experiment shows the potential improvement that can be obtained
in our generated datasets in terms of KL over the generated data.
For this we measure in Figure~\ref{fig:KLUnderlying} the KLs obtained 
by learning the parameters for three proposed structures: 
\textit{i)} the empty structure,
\textit{ii)} the fully connected structure,
and \textit{iii)} the underlying structure.
The distribution learned from the empty structure informs us about the impact of 
encoding incorrect independences in the KL measure. 
Consequently, the fully connected structure shows the impact in the KL measure that can be obtained 
with incorrect dependences that are obscuring the real CSIs present in data.
The underlying structure contains the features which exactly encodes the CSIs of the proposed model, 
as described in Section~\ref{sec:datasets}.
The figure shows the average and standard deviation over 
our $10$ generated datasets for training set sizes varying from 500 to 40000 (X-axis), 
for different domain sizes $6,7$ and $8$.
In order to better show differences among the KLs we show it in log scale.

\begin{figure*}[t]
  \centering
    \includegraphics[width=0.32\textwidth]{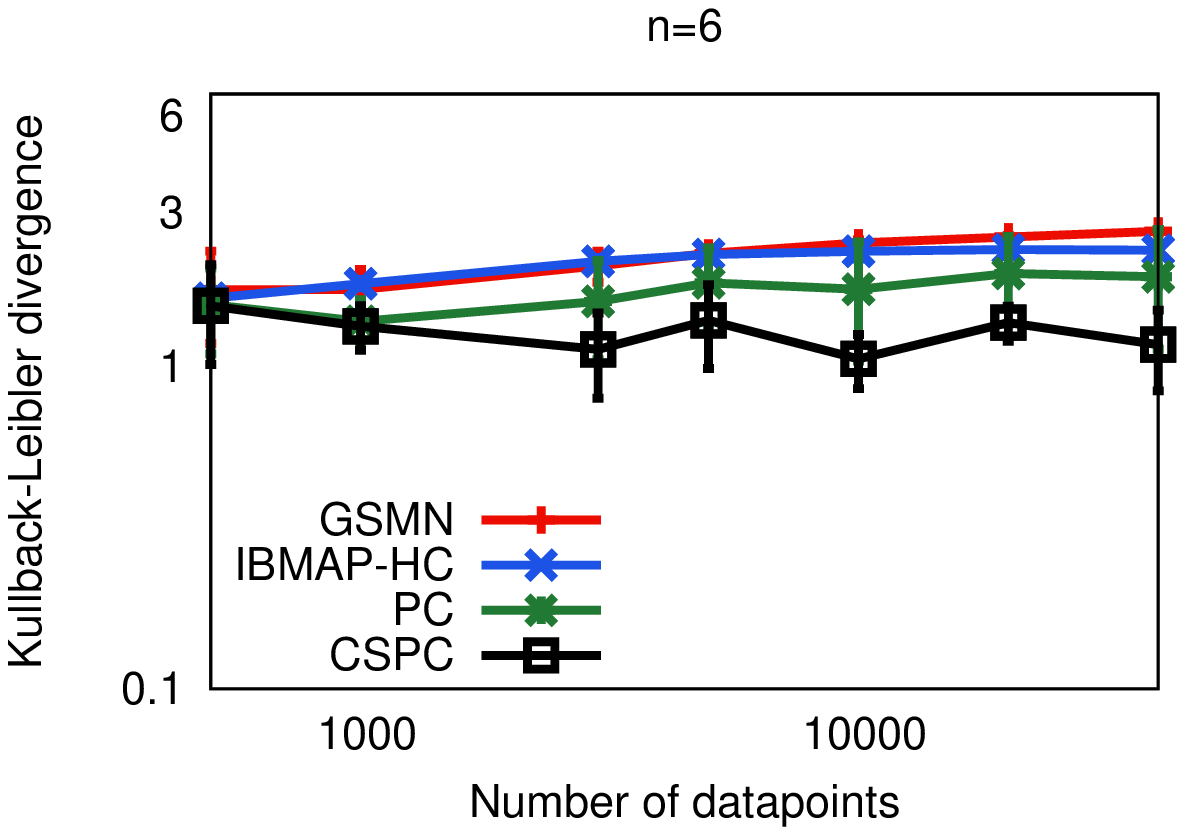}
    \includegraphics[width=0.32\textwidth]{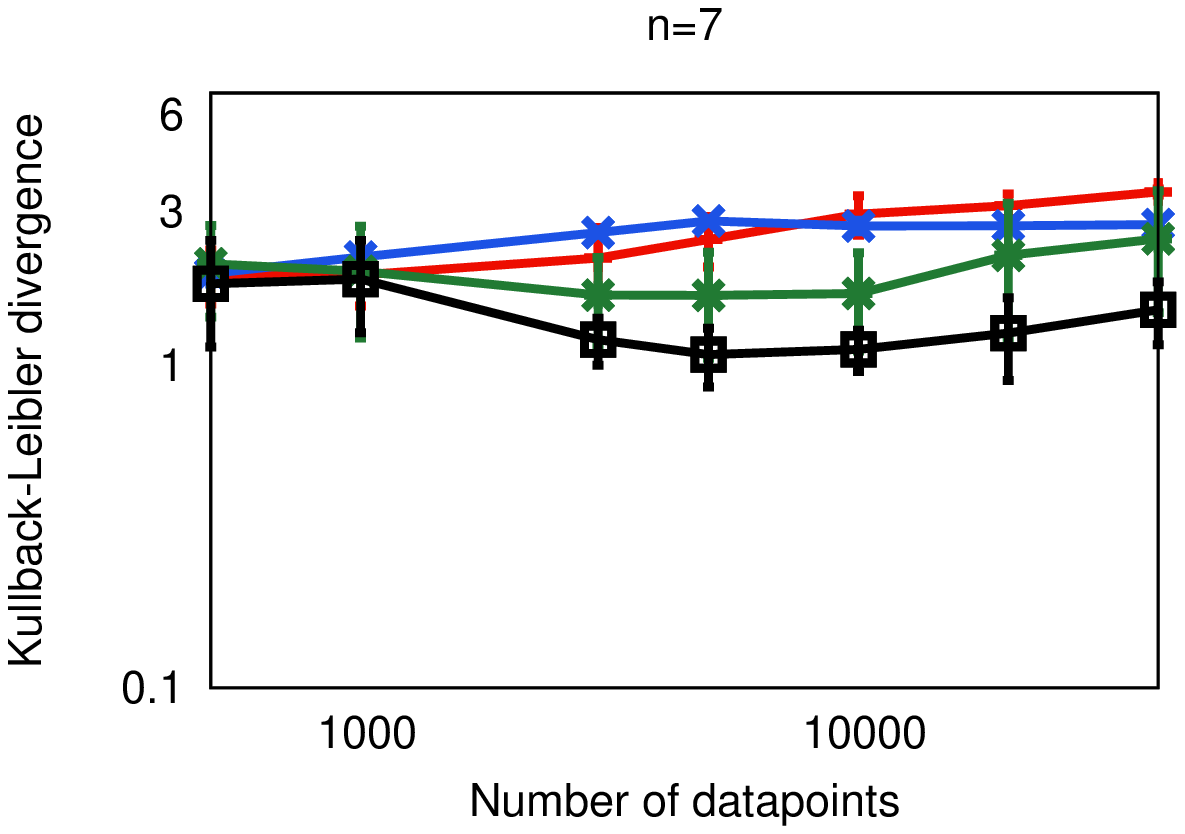} 
     \includegraphics[width=0.32\textwidth]{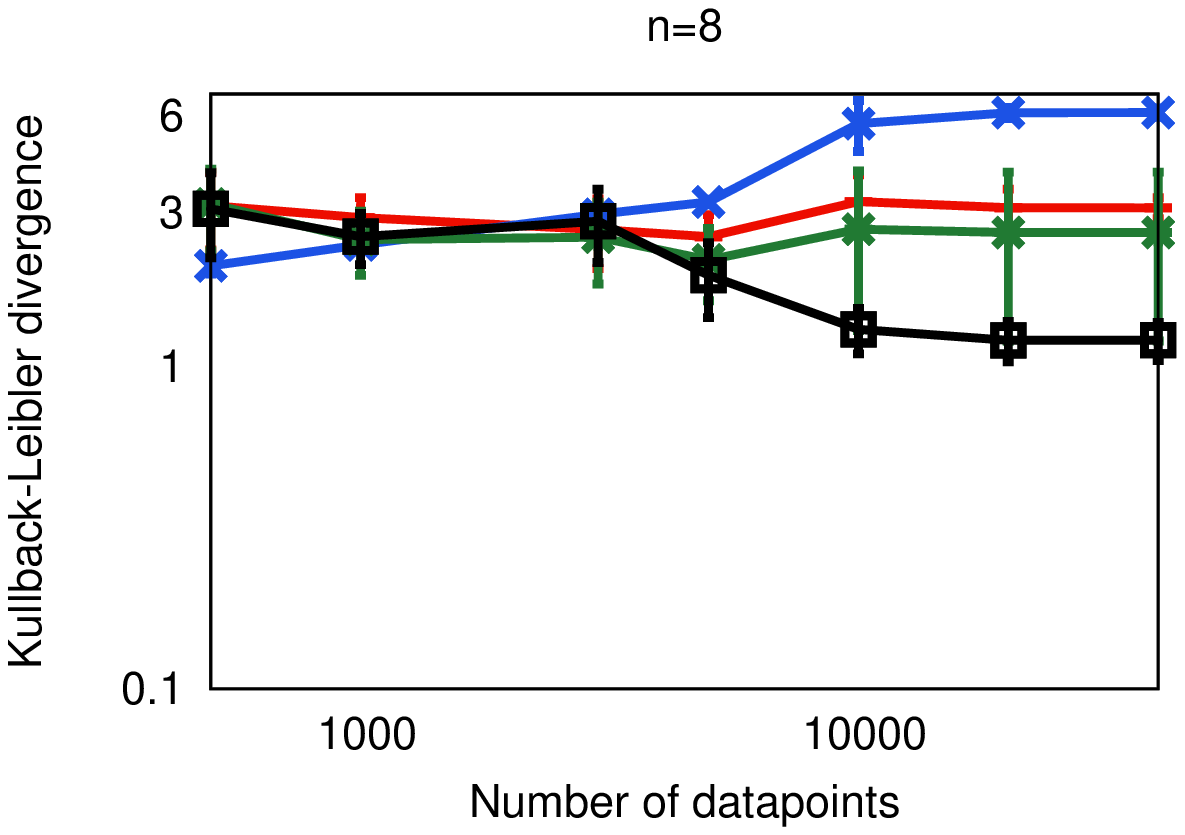}        
  \caption{
     Comparison of KLs obtained by learning parameters for CSPC, GSMN, IBMAP-HC and PC. 
     Average and standard deviation over ten repetitions for increasing number of datapoints 
     in the training set for domain sizes $6$ (left), $7$ (center) and $8$ (right). 
  }
  \label{fig:KLCompetitors}
\end{figure*}
\begin{figure*}[t]
  \centering
    \includegraphics[width=0.32\textwidth]{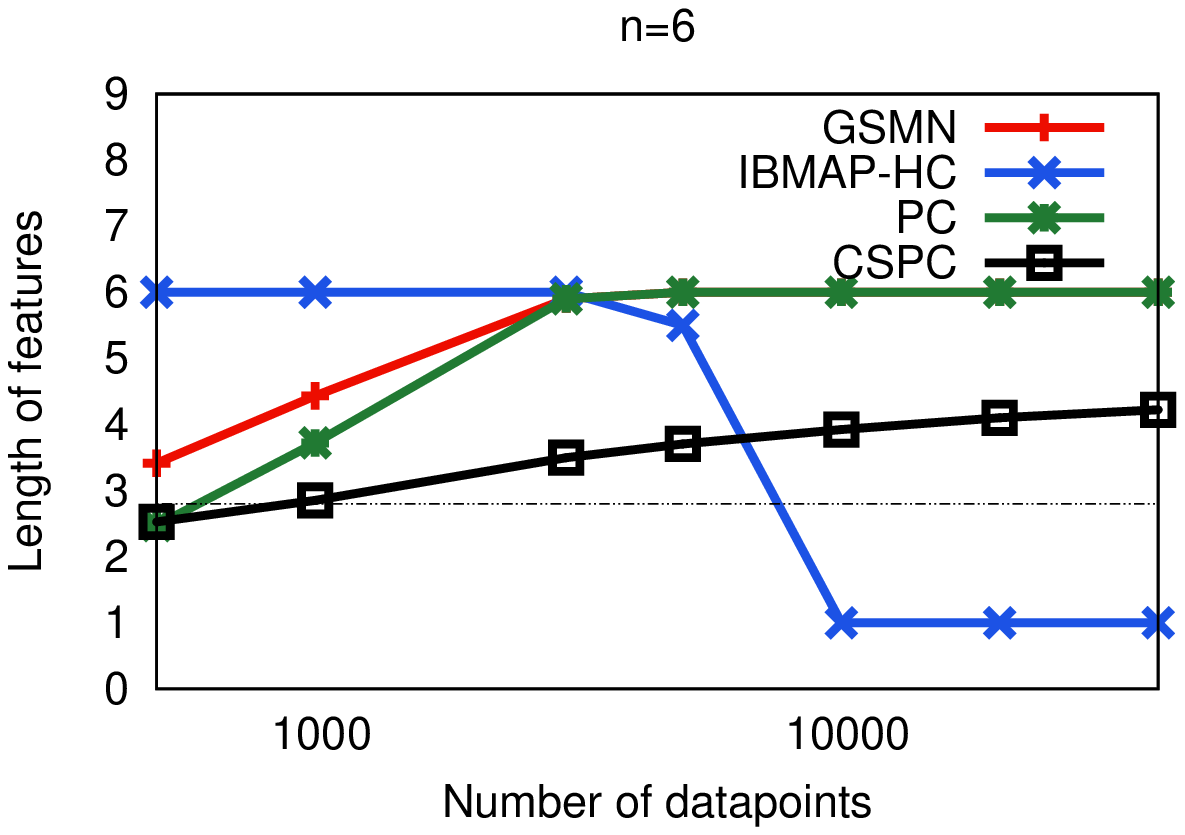}
    \includegraphics[width=0.32\textwidth]{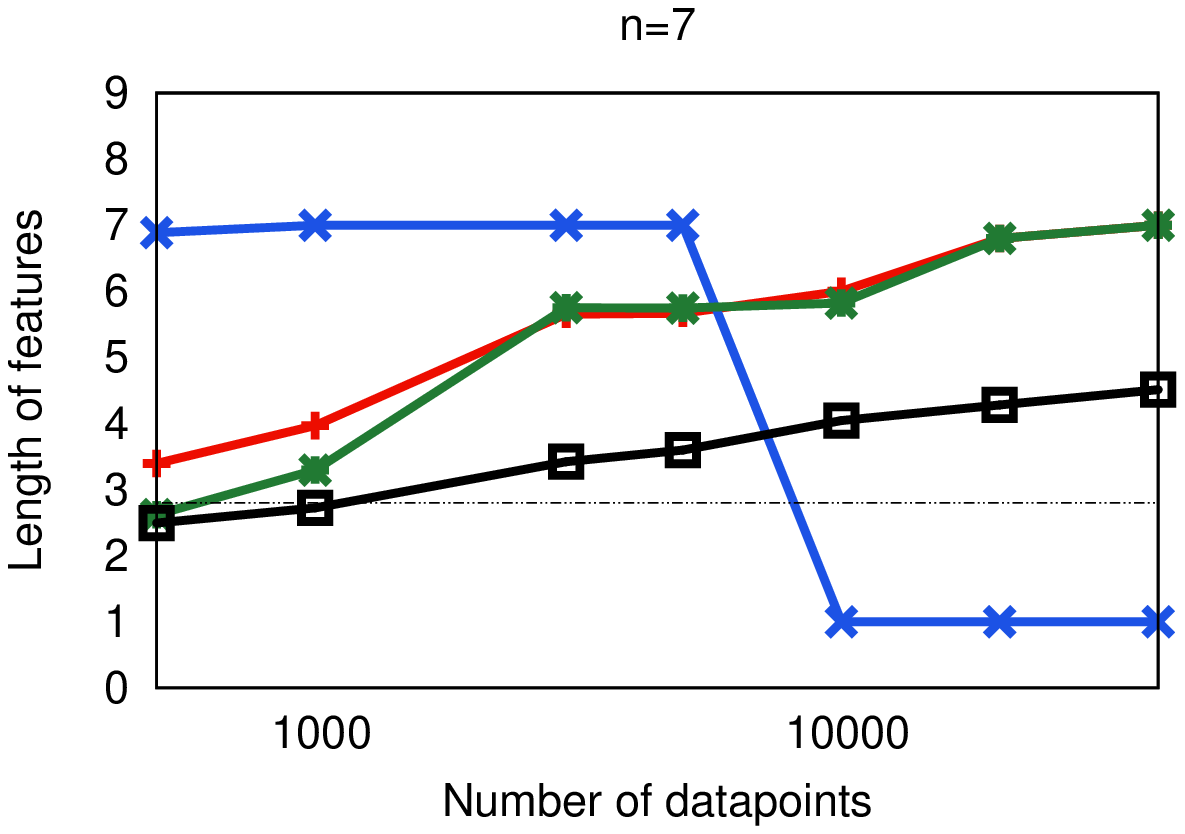} 
     \includegraphics[width=0.32\textwidth]{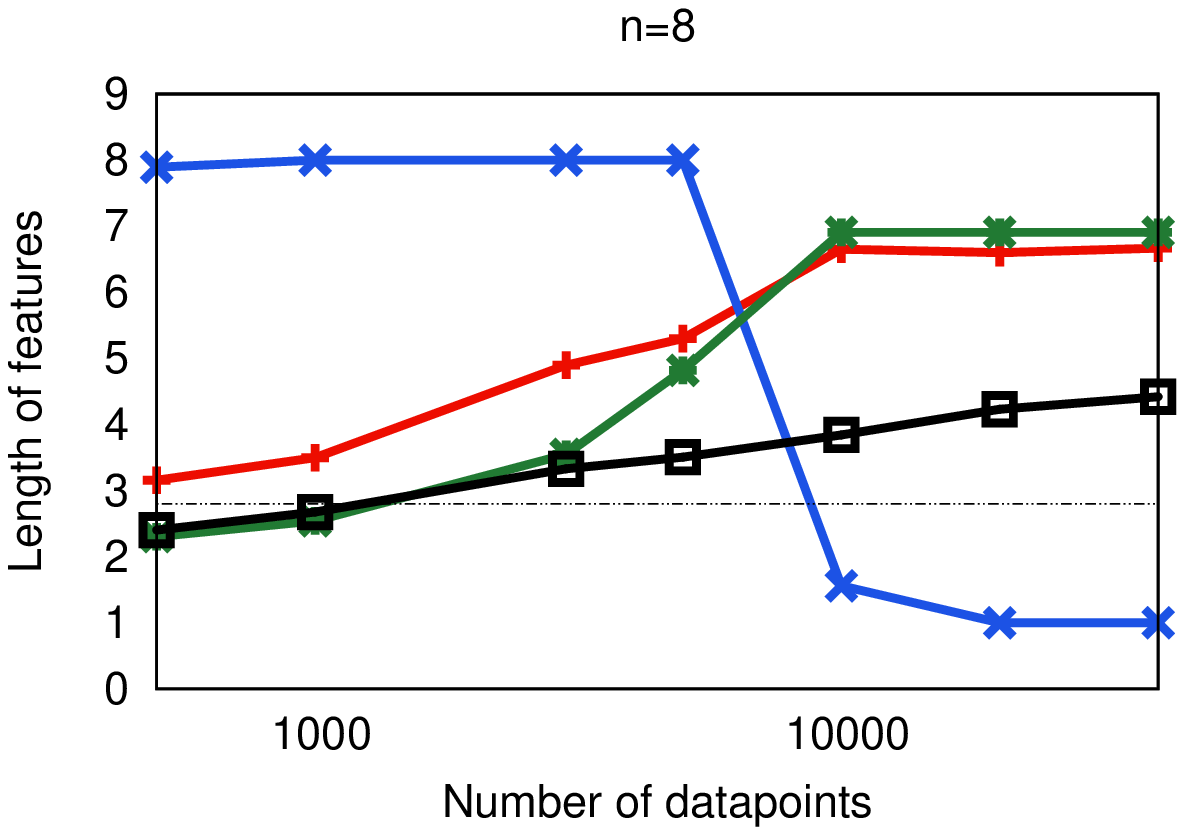}        
  \caption{
     Comparison of the average feature length obtained for CSPC, GSMN, IBMAP-HC and PC. 
     Average and standard deviation over ten repetitions for increasing number of datapoints 
     in the training set for domain sizes $6$ (left), $7$ (center) and $8$ (right). 
     The average feature length of the solution underlying structure is the horizontal line.
  }
  \label{fig:featuresLength}
\end{figure*}

In these results, we see empirically that the KL of the distribution obtained by learning the parameters
for the underlying structure is always significantly better than the KL 
obtained by using the empty and fully structures.
Notice that the KL is an (expected) logarithmic difference, representing differences in orders of magnitude 
in the non-logarithmic space. For our results, these differences are up to $2$ orders of magnitude
in the cases of $n \in \{6,7\}$, and $5$ orders of magnitude in the case of $n=8$.
However, there is not a trend of the KL to be zero by varying the size of training data.
This is because Gibbs sampler is not an exact method for the generation of the training data.
In this result also can be seen that as well as $n$ increases, 
the KL of the fully structure is better than the KL of the empty structure.

In our second experiment we compare the KLs obtained by CSPC, GSMN, IBMAP-HC, and PC.
These results are shown in Figure~\ref{fig:KLCompetitors},
for the same datasets of the experiment shown in Figure~\ref{fig:KLUnderlying}.
CSPC is the more accurate algorithm in all the cases, with lower KL,
near to $1$ (the KL value of the underlying structure in Figure~\ref{fig:KLUnderlying}).
The differences in KL between CSPC against its competitors is up to $2$ orders of magnitude for $n=6$ and $n=7$,
and up to $5$ orders of magnitude against IBMAP-HC for $n=8$.

Since the KL is clearly affected by the quality of the structure, we wanted to determine
whether or not their actual structures are correct.
We did this by reporting the average feature length of the learned models,
since it is a known value for our underlying model in this experiment.
This is a statistical measure useful for analyzing the structural quality of log-linear models,
as shown in several recent works \cite{lowd2010learning,van2012markov,lowd2013learning}.
Figure~\ref{fig:featuresLength} reports these values 
for the same experiment shown above.
The horizontal line in the graphs shows the exact feature length of the underlying structure 
($2.80$ for $n=6$, $2.83$ for $n=7$, and $2.85$ for $n=8$).
CSPC perform better in all the cases, showing always the nearest number of average feature length to the horizontal line.
This is consistent with the KL results shown in Figure~\ref{fig:KLCompetitors}.
GSMN and PC increases in the average number of features length as well as the number of datapoints grows, for all the domain sizes.
This trend is due because they are learning more dense structures as well as the number of datapoints grows, reaching the fully structure. 
For example, in the case of $n=7$, the GSMN and PC algorithms shows a trend to reach the fully structure,
and the KLs shown in Figure~\ref{fig:KLCompetitors} are similar to the KL of the fully structure shown in Figure~\ref{fig:KLUnderlying}.
Also, as well as $n$ increases, the difference on the average feature length between CSPC and its competitors
also increases. 
This is also consistent with the results shown in Figure~\ref{fig:KLUnderlying}. 
A surprising result is shown for IBMAP-HC, which does not show the same trend than GSMN and PC. 
It can be seen that for lower number of datapoints ($\mathcal{D}< 5000$) the algorithm
learns fully structures (the average feature length is equal to $n$ in the three cases).
However, for higher number of datapoints ($\mathcal{D} \geq 5000$) the algorithm learns the empty structure,
with average feature length equal to $1$ (the empty structure contains the atomic features). 
We argue this is due to the Bayesian nature of IBMAP-HC, which works by optimizing the posterior probability
of structures $p(G\mid \mathcal{D})$ with a hill-climbing search. 
When using a large amount of data, IBMAP-HC seems very prone to getting stuck 
in the empty structure as a local minima with $p(G\mid \mathcal{D})=0$ for almost all the structures, 
except the correct one with $p(G\mid \mathcal{D})=1$ .

Finally, as an additional result, we show in Table~\ref{table:numberfeatures} 
the average feature length of the features that are satisfied with the value of the flag variable $X_f$.
This result is shown for all the algorithms, 
running for an increasing number of variables from $n=4$ to $8$
with a fixed number of $3000$ datapoints, and discriminating in different columns the value of $X_f$.
As expected, the values for CSPC for $X_f=0$ are more near to $3$, and more near to $2$ for $X_f=1$,
in comparison with the rest of the competitors.
In summary, the above results support our theoretical claims and demonstrate the efficiency of CSPC
for learning distributions with CSIs.

\begin{table}[!ht] 
\center
\scriptsize
  \begin{tabular}{|m{.3cm}|m{.5cm}m{0.6cm}p{.5cm}p{.5cm}|m{.5cm}m{0.6cm}p{.5cm}p{.5cm}|}
    \hline
	
	& \multicolumn{4}{c}{$X_f=0$} &   \multicolumn{4}{|c|}{$X_f=1$} \\\hline
	$n$ & GSMN & IBMAP-HC & PC & CPSC & GSMN & IBMAP-HC & PC & CPSC   \\\hline
	4.00 & 4.00 & 4.00 & 4.00 & 4.00 & 4.00 & 4.00 & 4.00 & 1.85 \\
	5.00 & 5.00 & 4.60 & 5.00 & 4.79 & 5.00 & 4.60 & 5.00 & 1.88 \\
	6.00 & 6.00 & 5.50 & 6.00 & 4.37 & 6.00 & 5.50 & 6.00 & 1.93 \\
	7.00 & 5.00 & 7.00 & 1.10 & 4.01 & 5.00 & 7.00 & 1.10 & 1.95 \\
	8.00 & 3.00 & 8.00 & 1.00 & 3.54 & 3.70 & 8.00 & 1.00 & 1.87 \\ \hline
  \end{tabular}
  \caption{Number of features learned for increasing $n$, and using $\mathcal{D}=3000$.}\label{table:numberfeatures}
\end{table}

\section{Conclusions}
This paper proposed CSPC, an independence-based algorithm 
for learning a set of features, instead of a graph.
CSPC overcomes some of the inefficiency of traditional IB algorithms
by learning CSIs from data and representing them in a log-linear model.
CSPC proceeds by generalizing iteratively a set
of initial features in order to represent the CSIs present in data,
exploring the possible contexts, eliciting from data a set of CSIs usings statistical tests,
and generalizing the features according to the elicited CSIs.
Experiments in a synthetic case show that this approach is more accurate than
the state-of-the-art IB algorithms, when the underlying distribution contains CSIs.
Directions of future work include: adapting more efficient IB algorithms for learning CSIs; 
validation in real world datasets; comparison against state-of-the-art non-independence-based approaches
\cite{davis2010bottom,lowd2010learning,gogate2010learning,van2012markov};
adding Moore and Lee's AD-trees \cite{moore1998cached} for speeding up the execution of statistical tests, etc.

\bibliographystyle{IEEEtran}
\bibliography{cspc}

\end{document}